\documentclass{AISB2008}
\usepackage{times}
\usepackage{graphicx}
\usepackage{latexsym}

\begin{document}

\title{Impact of robot responsiveness and adult involvement on children's social behaviours in human-robot interaction}

\author{David Cameron\institute{Department of Psychology, University of Sheffield, UK, email: \{d.s.cameron, e.c.collins, a.millings, t.j.prescott\}@sheffield.ac.uk} , Samuel Fernando\institute{Department of Computer Science, university of Sheffield, UK email: \{s.fernando, r.k.moore, a.sharkey\}@sheffield.ac.uk} , Emily Collins$^1$, Abigail Millings$^1$, \\
Roger Moore$^2$, Amanda Sharkey$^2$, \and Tony Prescott$^1$ }

\maketitle
\bibliographystyle{AISB2008}

\begin{abstract}
A key challenge in developing engaging social robots is creating convincing, autonomous and responsive agents, which users perceive, and treat, as social beings. As a part of the collaborative project: Expressive Agents for Symbiotic Education and Learning (EASEL), this study examines the impact of autonomous response to children's speech, by the humanoid robot Zeno, on their interactions with it as a social entity. Results indicate that robot autonomy and adult assistance during HRI can substantially influence children's behaviour during interaction and their affect after. Children working with a fully-autonomous, responsive robot demonstrated greater physical activity following robot instruction than those working with a less responsive robot, which required adult assistance to interact with. During dialogue with the robot, children working with the fully-autonomous robot also looked towards the robot in anticipation of its vocalisations on more occasions. In contrast, a less responsive robot, requiring adult assistance to interact with, led to greater self-report positive affect and more occasions of children looking to the robot in response to its vocalisations. We discuss the broader implications of these findings in terms of anthropomorphism of social robots and in relation to the overall project strategy to further the understanding of how interactions with social robots could lead to task-appropriate symbiotic relationships. 
\end{abstract}

\section{INTRODUCTION}

A  key  challenge  for  human  robot  interaction  (HRI) research  is  the development of social robots that can successfully engage with human users. Effective  social  engagement  requires  robots  to  present  personalities promoting user interaction \cite{breazeal1999build} and to dynamically respond to, and shape their interactions to meet, user needs \cite{pitsch2009first}.

The Expressive Agents for Symbiotic Education and Learning (EASEL) project aims to develop a biologically grounded \cite{verschure2012distributed} robotic system that can meet these  requirements in the form of a socially-engaging Synthetic Tutoring Assistant (STA). Through developing the STA, we aim to take forward understanding of human-robot symbiotic interaction. Symbiosis in HRI, is considered to be the capacity of the robot and user to adapt to their partner and mutually influence interaction in a positive way \cite{charisitowards}. In a social HRI context, symbiosis requires that the robot can interpret and respond to user behavior or state, appropriately adapting its own actions. We draw from social psychological methods and models with the aim to uncover key factors in robot personality, behavior, and presentation that underpin symbiosis in HRI. We further hope that this work will shape a broader framework for exploring long-term and effective human-robot symbiotic interaction \cite{cameroninprep}.

The framework for affect-led human-robot symbiotic interaction \cite{cameroninprep} argues that during HRI with an unfamiliar social robot, users may not have established ideas of how to interact appropriately. As a result, they import their own social norms believed to be relevant, based on the robot's morphology and the interaction scenario (e.g., when in conversation with a humanoid robot, users will follow social conventions of turn-taking \cite{stivers2009universals} and maintaining appropriate eye-contact \cite{vertegaal2001eye} \cite{ito2004robots}). It is further argued that this process is more apparent with stronger user perceptions of a robot as a social entity \cite{cameroninprep}.

Understanding others as being social entities is a fundamental developmental process for children in their social cognition \cite{flavell1998social}. Social cognition provides a person their understanding of social situations faced, enabling them to make sense of ongoing social interactions and people in their social environment \cite{moskowitz2005social}. Elements of social cognition are seen early in infancy, including monitoring of others' actions and deriving meaning from others' gaze \cite{rochat2014early}. This typically develops with age and experience to include the understanding of others as having mental states, distinct from one's own \cite{birch2004understanding}. Children can apply their social understanding to interactions in HRI, to believe humanoid robots have mental states and can be social entities \cite{kahn2012robovie}. Although the factors influencing these perceptions and the impact these have on children's engagement in HRI with social robots are not fully understood.  

\begin{figure}
\centerline{\includegraphics[width=3.5in]{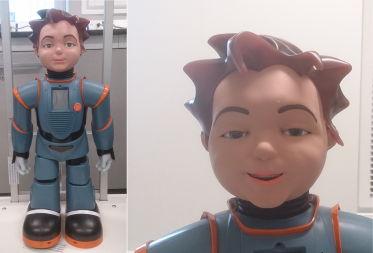}}
\caption{The Robokind Zeno R25 platform for the Synthetic Tutor Assistant (humanoid figure approximately 60cm tall)} \label{Zenofig}
\end{figure}

The STA model offers ideal means to explore the impact of users' perceptions of a robot as being a social entity on their interactions. The STA, presented through the Robokind Zeno R25 platform \cite{hanson2009zeno} (Figure \ref{Zenofig}), is developed to engage in collaborative inquiry-learning with children \cite{charisitowards}. Children perceive the Zeno R25 robot as being an animate machine-person hybrid, considered to be due to its humanoid appearance, responsiveness to the user, and autonomy in social interaction \cite{cameron2015children}. These anthropomorphic cues may give rise to children's perceptions of the robot as a social entity \cite{duffy2003anthropomorphism} \cite{belpaeme2013child} and so draw from their models of social cognition and applicable social norms. This paper explores the influence that the autonomy and responsiveness of a robot can have on children's behaviours relating to their perspectives of a robot being a social entity.

Important precursors to engagement with other social entities are attention coupling and eye contact \cite{argyle1965eye}; this is even seen in interaction with social robots \cite{kozima2003attention}. In conversation, eye contact can act as cues that individuals are listening and/or ready to listen \cite{vertegaal2001eye} and an early development in social cognition \cite{frischen2007gaze}. Importantly, a distinction could be made in the timing of looking to an agent: turning to look while another agent is currently talking may be regarded as a reaction to a sound stimulus; whereas looking in anticipation, indicates expectancy of an agent's response. Children can `converse' with inanimate objects (e.g., dolls), without aims of information exchange or even expectancy that they will respond \cite{gelman1981development} but expect response from animate social entities. Evidence of a user attributing social agency to a humanoid robot may therefore be seen in their \textit{anticipation} of the robot giving an answer to the user's posed question.

By recording participants (with parental consent), and through questionnaires,  we  obtained  measures of task engagement,  human emotional facial expression, gaze direction and reported affect. We hypothesized that children would interact with a robot autonomously responding to their speech by a) showing greater task engagement through more physical activity, b) reporting more positive affect and showing more positive expression during the interaction, and c) looking towards the robot more often in time for its turn in dialogue, compared to peers who interacted with the same robot that was responsive to adult speech only. Prior studies have shown some influence of demographics such as age and gender on HRI \cite{cameron2015presence}, \cite{kanda2004interactive}, \cite{shahid2010child}. In our study, a gender difference could also arise due to the Zeno robot being widely perceived as similar to a male child \cite{cameroninprep2}, which could prompt different responses in male and female children. We therefore considered these other factors as potential moderators of children’s experience of the interaction.

\section{METHOD}
The study took place as part of a voice data-collection exercise to assess and further calibrate the STA's automatic speech recognition system (ASR) \cite{fernandoinprep}. However, during the exercise, the ASR failed to detect some participants' voices. Fortunately, this enabled the present quasi-experiment; in which, children either interacted with the robot on their own or with adult assistance.
\subsection{Design} An independent measures design was used. Children were allocated to either the solo interaction or adult-assisted interaction conditions. Allocation was not random (i.e. determined by coin-flip) but determined by the ASR's capacity to recognise each child's voice in a brief interaction ahead of the main study (see section \ref{ASR Calibration}). 
\subsection{Participants}
The study took place at a local junior-school; children from a Year 4 class (ages 8 to 9) were invited to take part. Fourteen children completed the study (8 female, 6 male).
\subsection{Measures}
\subsubsection{Objective Measures}
The interaction included a series of brief physical activities for each child (e.g., jogging on the spot) that the system automatically tracked for an estimate of energy expended. Children's total energy (KiloJoules) expended is calculated.

Videos of participant expressions were recorded throughout the worksheet part of the interaction (see section \ref{Worksheet}) and automatically coded for discrete facial expressions: Neutral, Happy, Sad, Angry, Surprised, Scared, and Disgusted, using Noldus FaceReader Version 5. Mean intensity of the seven facial expressions across the duration of the game were calculated. FaceReader offers automated coding of expressions at an accuracy comparable to trained raters of expression \cite{lewinski2014automated}. 

The video recordings were further analysed to count the instances that participants looked towards a) the robot and b) the researcher during the interaction. Instances in which participants looked towards the robot were further divided into two subcategories: \textit{anticipatory} - looking towards the robot before it started answering the child's question - and \textit{reactive} - looking towards the robot after it had started answering the child's question.
\subsubsection{Questionnaire} \label{Questionnaire}
Participants completed the Self-Assessment Manikin (SAM \cite{bradley1994measuring}) for Valence, Arousal, and Dominance. Participants further completed a brief questionnaire on their enjoyment of the interaction and their beliefs about the extent to which they thought that the robot liked them, adapted from previous HRI work \cite{cameron2015presence}. Enjoyment of interacting with Zeno was recorded using a single-item, 100-point thermometer scale, ranging from `Really boring' to `Really enjoyable'. Participants' perceptions of Zeno's friendliness and the extent to which Zeno liked them were measured using single-item thermometer scales, ranging from `Not friendly at all' to `Very friendly' and `Not [liked me] very much’ to `Liked me a lot', respectively. Both these scales have been used  in our previous work exploring children's interactions with the Zeno Robot \cite{cameron2015children}
\subsection{Procedure}
The experiment took place in a local primary school, where participants completed the game under the supervision of the research staff and one member of school staff. Information regarding participation was sent before recruitment and informed consent was obtained from parents. Children were given full description of the tasks and then asked if they would like to take part. They were further informed that they could stop participating at any point without needing a reason and could ask the researchers any questions about the work.
\subsubsection{ASR Calibration} \label{ASR Calibration}
The interaction with Zeno began with the robot turning to face the participant and initiating dialogue with `Hello'. Participants then read the provided statement, `Hello Zeno are you ready to Start?'. If the ASR detected the participant's voice, Zeno would respond with, `Yes I am ready. I am Zeno the robot. I can understand simple words like Yes or No. Also I can understand anything that you read from the worksheet. Now please read the words underneath the picture of me'.

Participants then read `Testing A B C' which served as a further calibration phrase, and confirmation that the ASR was operating. If the phrase was correctly recognised the robot said `OK great. You can leave your worksheet on the table. Today I am going to help you learn about exercise and energy. First, you need to step on the mat so I can see you.'

If the ASR could not recognise the participant's voice, the robot said `Sorry I got that wrong. Please can you try again?'. If after two attempts the ASR still failed, the researcher would then assist by directing the participant to read five phrases for future calibration \cite{fernandoinprep} and then completing this part of the interaction on behalf of the participant. It was at this point in the study that participants were allocated to condition (solo or adult-assisted interaction).
\subsubsection{Physical Activity} \label{Exercise}
In all cases, participants moved on to the first task: three short sessions of physical activity, each directed by the robot. Participants were first instructed to move their arms for ten seconds; second, instructed to perform faster exercise for a further ten seconds (the robot offered examples such as jumping up and down or running on the spot); last, they were instructed to go a bit faster than before and for 20 seconds. All children completed this activity, as directed by the robot.

The three sessions were monitored using a Microsoft Kinect sensor; an estimation of the kinetic energy used was calculated and provided as feedback after each session. At the end of the interaction, the robot concluded by saying `OK well done. You completed 3 sessions. In total you used [number] kilojoules of energy. Now, when you are ready you can continue with the worksheet. Read out the questions and I will give you the answers.’
\subsubsection{Worksheet} \label{Worksheet}
Following the physical activity, participants completed the provided worksheet about their exercise and related questions on healthy living. The worksheet contained thirteen questions, which had been developed with primary-school teachers to meet the UK National Curriculum content for Year 4 science topics relating to healthy living. Example questions include, `How much energy did I use in the first exercise session?' and `How much energy is in an apple?'. The worksheet was headed `You can find out the answers by asking Zeno the Questions'.

Participants read out loud each of the questions provided, or, in the adult-assisted condition, the researcher read the questions on their behalf. The robot would then verbally respond with the answer for the child to write on the worksheet. Participants progressed through the worksheet and ended the interaction by reading the final statement `Thank You Zeno, goodbye'. Zeno would reply `Ok thank you for talking with me today. Goodbye!'.
\subsubsection{Post HRI}
At the close of the interaction, participants completed the brief questionnaire about their affect and perceptions of the activity and robot (as discussed in section \ref{Questionnaire}). Participants were free to ask any further questions about Zeno or discuss the activity.
\section{RESULTS}
A preliminary check was run to ensure even distribution of participants to condition. There were four female participants and four male participants in the solo condition and 4 female and two male participants in the adult-assisted condition. A chi square test run before main analysis to check for even gender distribution across conditions indicates no significant differences (x$^2$ (1, N = 14) = .39, \textit{p} = .53).
\subsection{Objective Measures} \label{objectiveres}
There was a significant main effect for condition on participants' degree of physical activity in the task \textit{F}(1,13) = 5.92, \textit{p} = .04. Participants in the solo interaction condition completed significantly more physical activity in comparison to those in the adult-assisted condition (M = 27.00kJ, SE = 6.60 versus M = 52.36kJ, SE = 8.08). This is a large observed effect (\textit{d} = 1.31)\footnote{For the standardised measure of effect size, Cohen's \textit{d}, the guidelines of small (.2) medium (.5) and large(.8) are used.}. There was no significant effect of child gender \textit{F}(1, 13) = .032, \textit{p} = .86, nor interaction effects between condition and gender \textit{F}(1, 13) = .18, \textit{p} = .68.

There were significant main effects for condition on the both the number of instances participants looked towards the robot in anticipation (\textit{F}(1,13) = 10.38, \textit{p} < .01) and in reaction (\textit{F}(1,13) = 6.32, \textit{p} = .03 ) to its speech. On average, participants looked towards the robot in anticipation on more occasions in the solo interaction condition in comparison to the adult-assisted condition (M = 13.67, SE = 1.33 versus M = 8.00, SE = 1.15). This is a large effect (\textit{d} = 1.75). In contrast, participants looked towards the robot in reaction on fewer occasions in the the solo interaction condition in comparison to the adult-assisted condition (M = 1.33, SE = .95 versus M = 4.50, SE = .83). Again, this is a large observed effect (\textit{d} = 1.30).

\begin{figure}
\centerline{\includegraphics[width=3.5in]{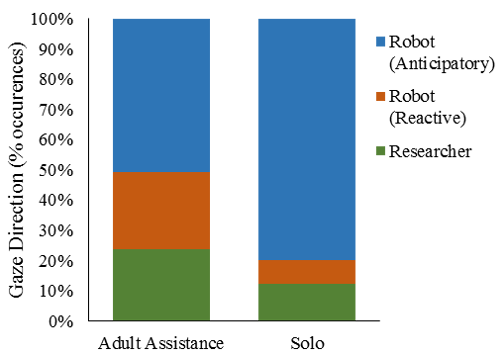}}
\caption{Percent occurrence of children's looking to figures, during worksheet phase of HRI} \label{Gazefig}
\end{figure}

There was no effect for condition on the total number of instances of looking towards the robot (\textit{F}(1,13) = .88, \textit{p} = .37), nor researcher (\textit{F}(1,13) = 3.00, \textit{p} = .11). When expressed as percentages of the the total instances of looking (anticipatory to robot, reactive to robot, and towards the researcher), results remain significant for both anticipatory (\textit{F}(1,13) = 13.03, \textit{p} $<$ .01) and reactive (\textit{F}(1,13) = 32.7, \textit{p} $<$ .01 ) looking towards the robot. On average, participants looked towards the robot in anticipation for a greater percent of occasions in the solo interaction condition in comparison to the adult-assisted condition (M = 79.87\%, SE = 6.11 versus M = 50.72\%, SE = 5.29). In contrast, participants looked towards the robot in reaction for a smaller percent of occasions in the the solo interaction condition in comparison to the adult-assisted condition (M = 7.74, SE = 2.36 versus M = 25.60, SE = 2.04). Figure \ref{Gazefig} highlights the impact of condition on both classifications of user looking.

There was a significant effect for condition on children's average expressions of sadness \textit{F}(1,13) = 6.74, \textit{p} = .03. Participants showed greater average sadness in the solo interaction condition in comparison to those in the adult-assisted condition (M = 5.91\%, SE = 2.12\% versus M = 14.63\%, SE 2.60\%). This is a large observed effect (\textit{d} = 1.40). There was no significant effect of child gender \textit{F}(1, 13) = 2.82, \textit{p} = .12, nor interaction effects between condition and gender \textit{F}(1, 13) = 1.88, \textit{p} = .20. There were no further significant effects for any of the remaining expressions.
\subsection{Questionnaire}
\begin{figure}
\centerline{\includegraphics[width=3.5in]{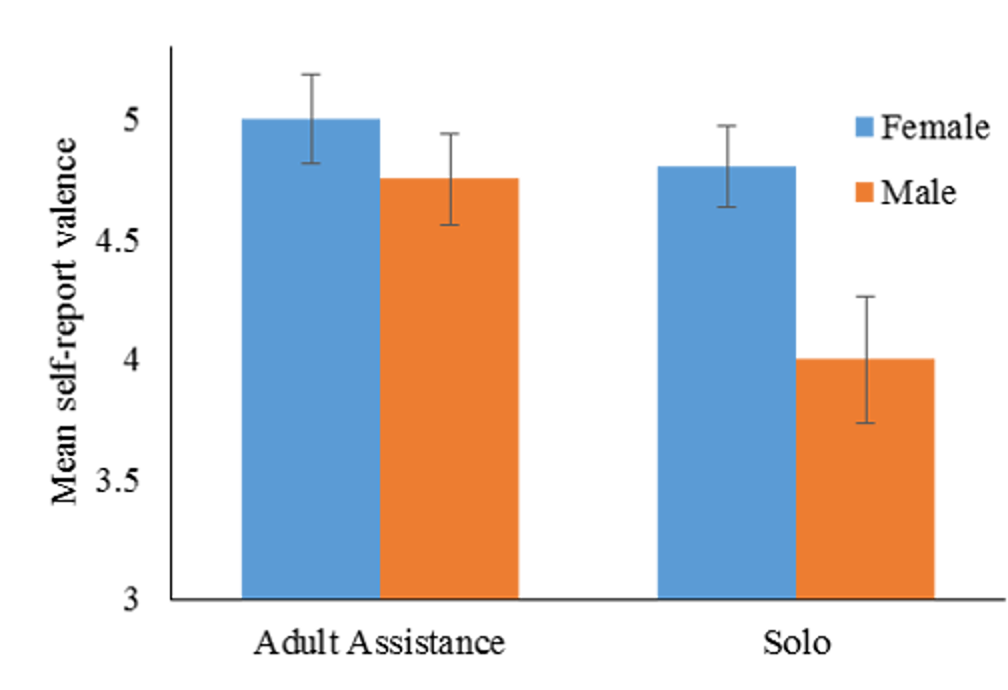}}
\caption{Mean ratings of post-HRI valence} \label{Valencefig}
\end{figure}

There were significant main effects on participants' self-reports of valence for both conditions \textit{F}(1,13) = 5.33, \textit{p} = .04 and gender \textit{F}(1,13) = 5.33, \textit{p} = .04. Participants reported greater average valence in the adult-assisted in comparison to the solo interaction condition (M = 4.88 SE = .14 versus M = 4.38 SE = .17). This was a large effect (\textit{d} = 1.25). On average girls reported being happier than boys following the interaction; there was no interaction effect observed. Results for condition and gender are presented in figure \ref{Valencefig}. There were no further effects found for the remaining SAM measures.

There were were no significant differences between conditions for children's ratings of their enjoyment of the interaction, their perceptions of Zeno as being friendly, nor the extent to which they think the robot may like them (Max \textit{F}(1,13) = 1.81, \textit{p} = .21).

\section{Discussion} 
The results provide new evidence that children's behaviour and affect in HRI can be influenced by a robot's responsiveness to the child. Our results show consistency across two behavioural measures and across two measures of affect. The robot's direct engagement with, and response to, the children could be seen to impact on behaviour, resulting in greater socially-relevant attention paid to the robot and engagement with the robot-directed task. However, affective measures do not indicate that children showed a more positive emotional experience in working with the robot directly in comparison to working with an adult during HRI. The behavioural and affective measures may relate to different aspects of the interaction, which we discuss in turn.

The current behavioural results could be due to children's perceptions of the robot as being an animate and social entity. In the solo interaction condition, children saw a fully-autonomous, responsive robot; in contrast, following ASR failure, the resetting of the robot and intervention by researchers may have robbed the machine of its apparent autonomy and responsiveness\footnote{The robot still exhibited response to the researcher's voice but the failure to respond to the child's voice \textit{in particular} may be sufficiently salient to a child}. In our prior work, children report the robot as being like a person \cite{cameron2015children} unless the robot is directly operated by a researcher \cite{cameroninprep2}. This anthropomorphic view is argued to impact on perceptions of robots as social entities \cite{duffy2003anthropomorphism} \cite{belpaeme2013child}. Children regarding the robot as a social entity may import their own relevant social norms about following instruction, and so be more likely to do so (in this case, exercise more; section \ref{objectiveres}). This effect may be particularly prominent, given the school setting and Zeno's introduction of itself as a personal trainer before giving instructions to exercise (section \ref{Exercise}). In contrast, children who perceive the robot as a machine rather than a social entity may not feel the same obligation to follow instruction because the social norms do not apply \cite{cameroninprep}.

The differences between conditions for children's looking towards the robot (Figure \ref{Gazefig}) further suggest differences in children's perceptions of the robot as a social entity. Children in the solo interaction condition tended to look towards the robot in anticipation of its answer to a substantially greater degree than those in the adult-assisted condition. Gelman \cite{gelman1981development} highlights the difference in expectancies children can have for animate and inanimate beings. While children can happily talk with inanimate objects, they do so without expecting a response; in contrast, conversation with another person predicts a response \cite{gelman1981development}. Children's looking towards the robot may indicate the same anticipation of response as expected when conversing with another person.

It is interesting to note that children's looking to the researcher did not vary in frequency between conditions. An adult mediating interaction between child and robot did not draw children's attention away from the interaction any more than an adult present, but `outside' of the HRI scenario. While not formally explored in the current work, children's looking towards the experimenter during the solo interaction condition typically coincided with the ASR misclassifying children's voices and so the robot giving incorrect or nonsensical answers to questions, potentially breaking social norms. Robots that break social norms have previously been demonstrated to be held to similar standards to humans \cite{snijders2015robot}. This may further indicate the impact of imported social norms on interaction: children expect the robot to adhere to their social norms and seek guidance when the robot appears to break them. 

The difference between conditions for children's reported valence and their recorded expressions may be related to the adult assistance rather than perceptions of the robot as a social entity. The findings may considered in terms of social reinforcement \cite{harris1964effects} and positive adult attention\cite{maag1999behavior}: working with an adult on a task in school may simply be more rewarding than otherwise. Alternatively, after working with an adult on a task, children may show expectancy effects, leading to more favourable ratings of tasks.

\section{Future Work}

Each of the findings are advised to be viewed with caution, given the small sample size; larger-scale replications are vital in better establishing the impact responsiveness can have on children's social behaviours. These results indicate potentially fruitful topics of research in HRI and we invite researchers to explore these. The study design arose fortuitously out of unfortunate circumstances; further experimental work developed to target particular elements of this quasi-experiment - autonomy, reliability, researcher involvement in the HRI activity - could better identify which, if any, influence the HRI experience.

We aim to repeat the current study with two key developments: 1) use stricter experimental procedure and controls and 2) make use of measures regarding animacy and social agency not available to the research team during the ASR piloting. 

First, rather than allocate condition by ASR success or failure, random allocation would be preferred (ASR would be randomly tuned to either child- or adult- voice recognition). With random allocation, a third condition would be added of \textit{adult-led interaction}. This enables a subtle but meaningful revision to procedure: children in the adult-led condition do not first see the robot falter in the task; it is simply stated that the adult will read the worksheet. Inclusion of this condition can explore the impact of perceived reliability on the children's behaviour. In the present research, allocation to the adult-assisted condition necessarily coincided with children observing an apparent faulty robot. Perceived reliability of a robot can impact on user engagement with instruction \cite{salem2015would} and may contribute to current findings. 

Second, measures of animacy from the Godspeed questionnaire \cite{bartneck2009measurement} and our past work \cite{cameron2015children} would be used and be anticipated to correlate with our behvioural indexes reported in this paper. Open-ended questions about the robot's status as a social agent, inspired from \cite{kahn2012robovie}, could further indicate the impact autonomous responses to the user have on perceptions of the robot as a social entity. Again, we would expect use of words or phrases surrounding the concepts of the robot being animate and a social entity to coincide with our behavioural measures treating it so.

\section{Conclusion}
This paper offers further steps towards developing a theoretical understanding  of symbiotic interactions between humans and robots. The influence of a robot's responsiveness to users is identified as a factor in shaping human perceptions of a robot as a social being and, in turn, behavioural differences during HRI; follow-up work to examine this is identified. These findings highlight important considerations to be made in future developments of socially engaging robots.

\ack
This work is supported by the European Union Seventh Framework Programme (FP7-ICT-2013-10) under grant agreement no. 611971. We wish to acknowledge the contribution of all project partners to the ideas investigated in this study.

\bibliography{aisb}

\begin{thebibliography}{10}
\bibitem{cameroninprep2}
D~Cameron, S~Fernando, A~Millings, K~Moore, and T~Prescott, `You made him be
  alive! young children’s beliefs of a humanoid robot’s animacy are lower
  when observing direct operation', (Manuscript in Preparation).
\end{thebibliography}

\end{document}